\documentclass{cta-author}

{}
{}
{}

\begin{document}

\supertitle{This paper is a preprint of a paper accepted by IET Computer Vision and is subject to Institution of Engineering and Technology Copyright. When the final version is published, the copy of record will be available at the IET Digital Library.}

\title{TanhExp: A Smooth Activation Function with High Convergence Speed for Lightweight Neural Networks}

\author{\au{Xinyu Liu$^{1, 2}$}, \au{Xiaoguang Di$^{1\corr}$}}

\address{\add{1}{Control and Simulation Center, Harbin Institute of Technology, Harbin 150080, P.R. China}
\add{2}{Department of Electrical Engineering, City University of Hong Kong, Hong Kong SAR, China}
% \add{3}{Third Department, Third University, Address, Country Name}
% \add{4}{Current affiliation: Fourth Department, Fourth University, Address, Country Name}
\email{dixiaoguang@hit.edu.cn}}

\begin{abstract}
Lightweight or mobile neural networks used for real-time computer vision tasks contain fewer parameters than normal networks, which lead to a constrained performance. In this work, we proposed a novel activation function named Tanh Exponential Activation Function (TanhExp) which can improve the performance for these networks on image classification task significantly. The definition of TanhExp is $f(x)=x\tanh(e^x)$. We demonstrate the simplicity, efficiency, and robustness of TanhExp on various datasets and network models and TanhExp outperforms its counterparts in both convergence speed and accuracy. Its behaviour also remains stable even with noise added and dataset altered. We show that without increasing the size of the network, the capacity of lightweight neural networks can be enhanced by TanhExp with only a few training epochs and no extra parameters added.
\end{abstract}

\maketitle

\section{Introduction}\label{sec1}

Lightweight neural networks, also known as mobile neural networks, are specially designed for realizing real-time visual information processing. They tune deep neural network architectures to strike an optimal balance between accuracy and performance, meanwhile tailored for mobile and resource limitted environments [1]. These networks are necessary for computer vision tasks which require real-time computation [2-5]. These scenarios bring a big challenge because they restrict the size of the model and the training time. Therefore, the shallow structures and few layers with trainable parameters of these networks constrain their ability to simulate a non-linear function precisely. Noticing that the powerful ability of a neural network to fit a non-linear function lays upon the activation function inside, we consider that an effective activation function can help boost the performance of these networks without sacrificing size and rapidity.

Previous researchers mainly aim at exploring the best design of the activation function for normal neural networks. From the initially used Sigmoid to the recent Mish [6], researchers have proposed a great number of activation functions. Among of them, the most widely-used is the Rectified Linear Unit (ReLU) [7] because it is computed straightforward meanwhile shows an acceptable performance. Lightweight neural networks also adopt ReLU as the activation function because of the advantages elaborated above. However, with a non-zero mean, ReLU suffers from a bias shift problem. Each unit with ReLU activated will cause a slight bias shift, thus a series of units will make the situation severe. Besides, a mean far from zero decelerates the learning speed as well. Therefore, it cannot fully develop the efficiency of lightweight neural networks. But designing an adequate activation function for them was overlooked by previous researchers, while none of the other activation functions proposed by researchers could replace the practical and simple ReLU in these networks at present because most of them are complex while the improvement is negligible. Besides, these functions are not robust for the variation of data and the addition of noise. 

In this work, we propose a Tanh Exponential Activation Function (TanhExp), which combines the advantages of activation functions similar to ReLU and other non-piecewise activation functions together. Meanwhile, it requires little time for computation, which is suitable for lightweight neural networks. TanhExp is a continuous function with negative values and the positive part is approximately linear. These properties of the TanhExp accelerate the training process meanwhile ensure the sparsity of the input data. We demonstrate the efficiency, simplicity, and robustness on various datasets and networks, TanhExp shows much more noteworthy improvement than its counterparts. 

The paper is organized as follows. Section 2 introduces the related works. In Section 3 we give a detailed description of our TanhExp, which includes its definition, derivatives, and properties. In section 4, we demonstrate the simplicity and efficiency of TanExp on several datasets and show the results. Section 5 gives the conclusion of the whole work.

\begin{figure}[!b]
\begin{center}
\includegraphics[width=0.7\linewidth]{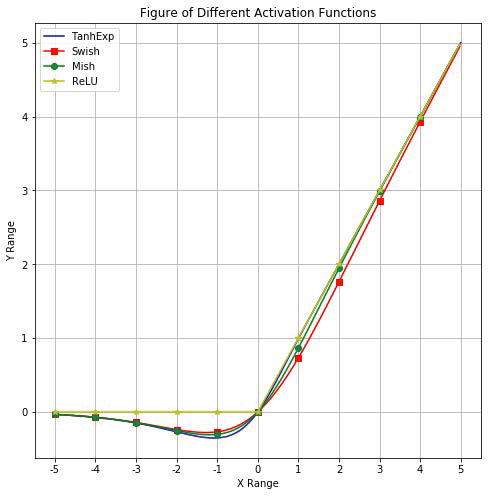}
\end{center}
\caption{TanhExp, Mish, Swish and ReLU Activation Function. We restrict the x coordinate from -5 to 5 for a clear view.\label{fig1}}
\end{figure}

\section{Related Work}\label{sec2}
\subsection{The ReLU Family}
On the initial stage of designing deep neural networks, Sigmoid and Tanh were commonly utilized as the activation function due to their non-linearity. However, the saturation of these two functions could severely restrict the fitting ability of the network and may cause a gradient vanishing. Therefore, Rectified linear units, i.e. ReLU [7] was proposed as a new type of activation function, which was defined as \(f(x) = max(0,x) \). Different from the previous functions, ReLU does not saturate on the positive half, thus it has two advantages: avoiding the gradients from vanishing and accelerating the learning speed. Although ReLU has been widely used, people still doubt whether ReLU is the best solution for all circumstances. Later researchers found out that ReLU has several drawbacks. The first is that ReLU is a non-negative activation function, thus it has a mean value above zero, which may cause a bias for the network layers afterward. Therefore, the deeper the network, the larger the bias. Besides, a zero-mean activation function can bring the gradient closer to the natural gradient thus accelerates the learning process [8], while ReLU does not have the ability. The second is the hard truncation of ReLU, which refers to the complete zero in the negative part. If a large gradient flows into ReLU, it will show no activation to the latter data, which was named as the 'Dying-ReLU' problem.

In order to overcome these drawbacks, researchers came up with several ideas. Leaky Rectified Linear Unit (Leaky ReLU) [9] \(f(x) = max(0,x) + leak\cdot min(0,x) \) adds a small slope in the negative part, where \(leak\) is a constant defined before training. Parametric Rectified Linear Unit (PReLU) [10] gives a similar solution as Leaky ReLU, with the slope rate in the negative part learned through data. However, it leads to a cost of learning extra parameters. S-shaped Rectified Linear Unit (SReLU) [11], which consists of three piecewise linear functions with the inflect point and the slope rate learned, suffers from additional parameters as well. Randomized Leaky Rectified Linear Unit (RReLU) [12] also uses \(max(0,x)\) as the positive part, but the negative part was replaced by a randomized Leaky ReLU. Exponential Linear Unit (ELU) [8], defined as \(f(x) = max(0,x) + min(e^x-1,0) \), uses the exponential function to generate a more smooth activation function. Scaled Exponential Linear Unit (SELU) [13] is a modified version of ELU, defined as \(f(x) = \lambda(max(0,x) + min(\alpha(e^x-1),0)) \) where \(\lambda \approx 1.0507, \alpha \approx 1.6732\), which was derived from a mathematical deduction. Gaussian Error Linear Unit (GELU) [14] was utilized in Bidirectional Encoder Representations from Transformers (BERT) [15], it combines properties from dropout, zoneout, and ReLUs. These methods tend to design a piecewise function with a smooth figure and force its mean close to zero. Nevertheless, the negative part of some of these activation functions loses the sparsity of ReLU [7]. Besides, the improvement is trivial but with more parameters added. As a result, none of them could be widely used like ReLU. People are more tend to use the traditional ReLU rather than a more complicated function with additional parameters.

\begin{figure}[!b]
\begin{center}
\includegraphics[width=0.7\linewidth]{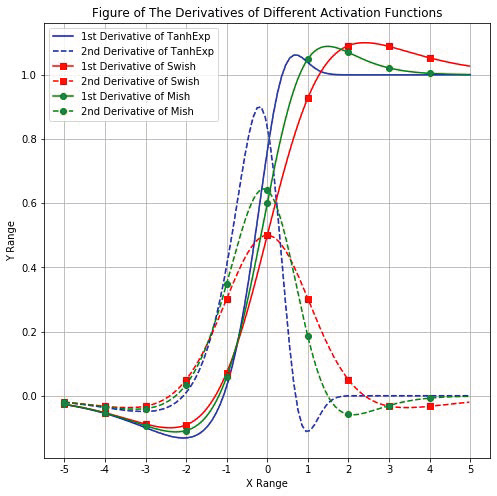}
\end{center}
\caption{The first and second derivatives of TanhExp, Swish, and Mish. \label{derivatives}}
\end{figure}

\subsection{Non-piecewise Activation Functions}

As mentioned above, Sigmoid and Tanh were initially used as activation functions in neural networks, but their saturation on infinity restricts their performance, i.e. a gradient vanishing problem. However, the ReLU family overcame the saturation problem, while led to other drawbacks. Therefore, is there an activation function that can meet all the above requirements? Swish [16] proposed a novel solution to design the activation function. It took advantage of an automated search technique to obtain activation functions, with a search space containing unary and binary functions. The experimental results indicate that \(f(x) = x\sigma(\beta x) \) outperforms all other counterparts on several tasks, which was named as Swish, where \(\sigma\) refers to the Sigmoid function in Eq. (\ref{equsigmoid}) and $\beta$ is a weight parameter.
\begin{align}
\mathbf \mathit{\sigma}(x) = \frac{1}{1+e^{-x}}
\label{equsigmoid}
\end{align}

Inspired by Swish, Mish [6] proposed a similar solution, its definition is $f(x) = x\tanh(\ln(1+e^x))$. Mish also provided many detailed experiments to demonstrate its superiority. Consequently, these activation functions not only inherit the advantages of ReLU but also bring about some other virtues. In detail, these functions are non-linear, which constructs the basic non-linearity of a deep neural network. They are unsaturated above, which could avoid the gradient vanishing problem. Soft-saturated at the negative infinity, which brings the sparsity to the network. Approximately being zero-mean, as elaborated above, a zero-mean function will lead the gradient closer to the natural gradient and accelerate the learning process. 

However, the research does not actually halt since the current activation functions are still not perfect, with the following problems existing. The first is a high computational complexity. For instance, the first derivative of Mish [6] can be calculated in Eq. (\ref{equmishhat}). It is complicated and slows down the backpropagation process critically. We will prove it in the experiment section. 

\begin{equation}
\label{equmishhat}
    f^{\prime}_{Mish}(x) = \frac{e^x(4(x+1)+4e^{2x}+e^{3x}+e^{x}(4x+6))}{(2e^x+e^{2x}+2)^2}
\end{equation}

The second is the introduction of parameters. Once a hyper-parameter is introduced in a network, the performance varies as the hyper-parameter varies, which cannot obtain a general solution to all tasks. Besides, if the parameter is trainable, it will definitely enlarge the size of the network, especially in a lightweight neural network. The third is that the previous methods overlooked the positive part. Swish and Mish are not approximately linear in the positive part, which will disturb the original distribution of the input data. Therefore, it leads to our proposed Tanh Exponential Activation Function, which can be abbreviated as TanhExp. Different from Swish and Mish, TanhExp generates a steeper gradient, alleviates the bias shift better, and preserves the distribution of the input.

\begin{figure}[!b]
\begin{center}
\includegraphics[width=1.0\linewidth]{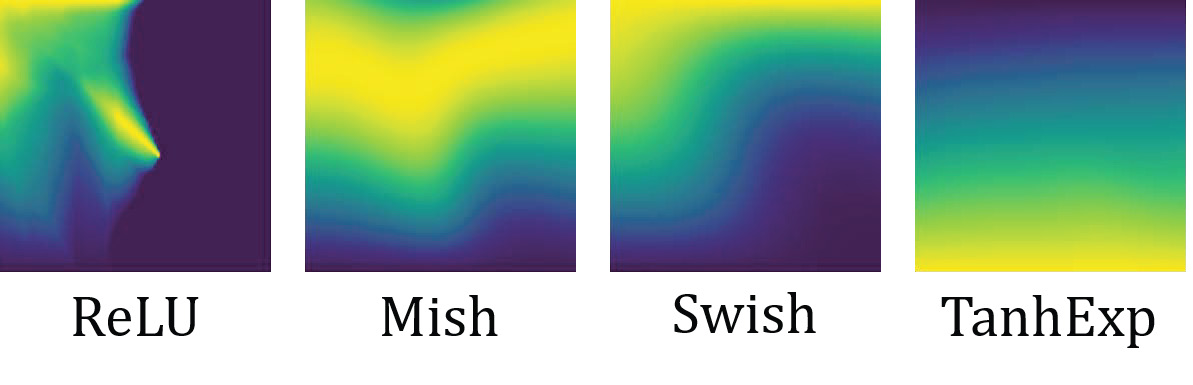}
\end{center}
\caption{Landscapes on a 5-layer network with different activation functions.\label{fig3}}
\end{figure}

\section{Tanh Exponential Activation Function}\label{sec3}

In this section, we introduce the Tanh Exponential Activation Function(TanhExp), which can be defined in Eq. (\ref{equtanhexp}):
\begin{equation}
    \label{equtanhexp}
    {f}(x) = x\tanh(e^x)
\end{equation}
where tanh refers to the hyperbolic tangent function:
\begin{equation}
    {tanh}(x) = \frac{e^x-e^{-x}}{e^x+e^{-x}}
\end{equation}

In the subsections below, we give a detailed description of the TanhExp and illustrate the properties.

\subsection{Graph and Derivatives of TanhExp}

The graph of TanhExp is shown in Fig. \ref{fig1}. Similar to other smooth functions, TanhExp also extends below zero at the negative part but has a larger gradient. The figure of TanhExp is intuitively similar to the figures of Mish and Swish, while they are not actually the same. Although they are close to each other, TanhExp requires less calculations. The first derivative of TanhExp can be calculated in Eq. (\ref{tanhexpderivative}).

\begin{equation}\label{tanhexpderivative}
\begin{aligned}
    {f^{\prime}_{TanhExp}}(x) &= \tanh(e^x) + x e^x {\rm sech}^2(e^x) \\ 
    &=\tanh(e^x) - x e^x(\tanh^2(e^x) - 1)
\end{aligned}
\end{equation}
The first and second derivatives of TanhExp, Swish, and Mish are shown in Fig. \ref{derivatives}.

\subsection{Properties of TanhExp}

TanhExp has a minimum value near $x=-1.100$, which is approximately $-0.3532$. TanhExp also inherits the so-called 'Self-gated' property defined in Swish [16]. The 'Self-gated' refers to a function with the form of $f(x) = x g(x)$. It is a multiply of the input itself and a function with the input as its argument, so the network will not change the initial distribution of the input on the positive part, meanwhile generates a buffer at the negative part near zero. TanhExp also ensures a sparsity of its output. According to [17], a sparse activation means in a randomly initialized network, not all inputs are activated. From the definition and the figure of TanhExp, we have
\begin{equation}
    \lim_{x \to -\infty}f_{TanhExp}(x) = 0
\end{equation}
Therefore, the neuron can be approximately treated as not activated when the input $x$ has a large negative value, which satisfies the definition of sparsity. This sparse property allows a model to control the effective dimensionality of the representation for an input, meanwhile more likely to be linearly separable. When compared with ReLU which suppresses 50\% of the hidden units, TanhExp has a smaller probability of deactivating these neurons. We consider that the noise in the data only accounts for a small proportion and ReLU will prohibit more useful features than TanhExp. Besides, a network with ReLU activated can be inefficient due to half of the neurons are not involved.

Although TanhExp seems similar to other smooth activation functions, it has several advantages over other smooth functions. 

Firstly, in the positive part, TanhExp is almost equal to a linear transformation once the input is larger than 1, with the output value and input value no more than 0.01 variation. As mentioned above, the ReLU family are all aiming at modifying the negative part while leaving the positive part its initial form. It is because the linear transformation is reasonable in training, yet the previous non-piecewise smooth activation functions ignored this property.

Secondly, TanhExp shows a steeper gradient near zero that can accelerates the update of the parameters in the network. During backpropagation, the network updates its parameters as Eq. (\ref{update}) shows. 
\begin{equation}
    w_{new} = w_{old} - \eta\triangledown w
    \label{update}
\end{equation}
Where $\eta$ refers to the current learning rate and $\triangledown w$ refers to the backpropagation gradient. $w_{old}$ and $w_{new}$ represent the weights of the network before and after updating, respectively.

We define $L$ as the loss of the network, which is a numerical representation of the differences between the network output and the ground truth label. For an image recognition task, one can use the cross-entropy loss as $L$. The cross-entropy loss is calculated in Eq. (\ref{celoss}).
\begin{equation}
    L=-\sum_{i=1}^{N} y^{(i)} \log \hat{y}^{(i)}+\left(1-y^{(i)}\right) \log \left(1-\hat{y}^{(i)}\right)
    \label{celoss}
\end{equation}
In the loss function, $N$ is the total number of samples, $y^{(i)}$ is the ground truth label for the $i-th$ sample and $\hat{y}^{(i)}$ refers to the network prediction for the $i-th$ sample. To improve the accuracy, the value of $L$ should be minimized by updating the parameters of the network. Then, based on the calculated loss $L$, $\triangledown w$ can be calculated as
\begin{equation}
    \triangledown w = \frac{\partial L}{\partial w_{old}}
\end{equation}
Therefore, if $\triangledown w$ is slightly larger, the update speed of the weight would be accelerated, thus leads to fast convergence. However, as our goal is to reach the global minimum value, an activation function with too large gradient might cause the network not to converge, while an approximately linear function is a rational option. Besides, from the theorem proved in ELU, we know that the bias shift of ReLU activated unit leads to oscillations and impede learning and the unit natural gradient can mitigate the problem. Moreover, a bias shift correction of the unit natural gradient is equivalent to shifting the incoming units towards zero and scaling up the bias unit. So the steeper gradient of TanhExp can also help to push the mean value of the function to zero, which further speeds up the learning process.

We also visualized a simple 5-layer fully connected network built with different activation functions in Fig. \ref{fig3}. Compared with ReLU, the other three activation functions shows a smoother landscape which indicates that Swish, Mish, and TanhExp avoid sharp transitions as ReLU does. Among these three smooth functions, TanhExp shows an especially continuous and fluent transition shape. This property guarantees that TanhExp is able to synthesize the advantages of both piecewise and non-piecewise activation functions and leads to outstanding performance. 

\begin{figure}[t]
\begin{center}
\includegraphics[width=1.0\linewidth]{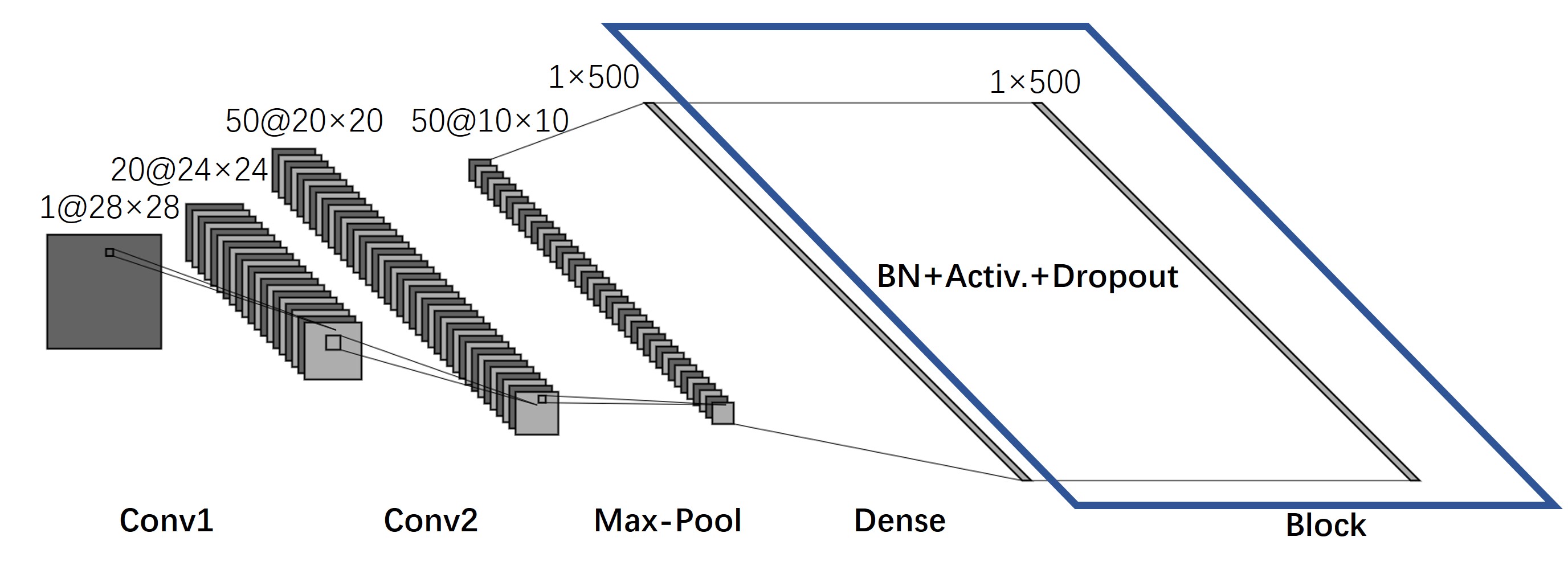}
\end{center}
   \caption{The network in the experiments on MNIST. Note that while tuning the layers of the network, the 'block' we duplicate refers to the part in the dark blue parallelogram. Depth, height, width are displayed in the 'Depth@Height$\times$Width' format. 'Activ.' refers to the activation function.}
\label{mnistnet}
\end{figure}

\begin{figure}[t]
\begin{center}
\includegraphics[width=0.7\linewidth]{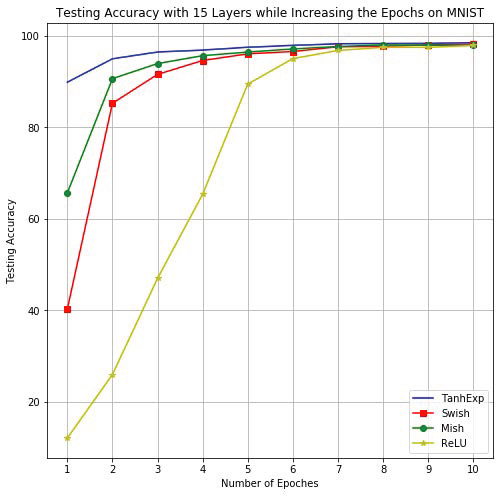}
\end{center}
   \caption{Testing accuracy (in percentage) with 15 layers on MNIST with different activation functions.}
\label{mnistacc}
\end{figure}

\begin{figure}[t]
\begin{center}
\includegraphics[width=0.7\linewidth]{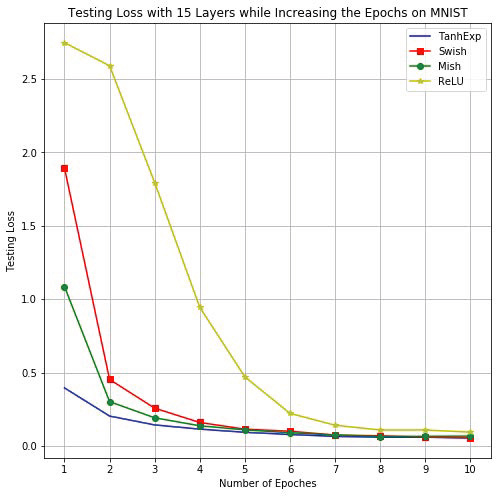}
\end{center}
   \caption{Testing loss with 15 layers on MNIST with different activation functions.}
\label{mnistloss}
\end{figure}

\begin{figure}[t]
\begin{center}
\includegraphics[width=0.7\linewidth]{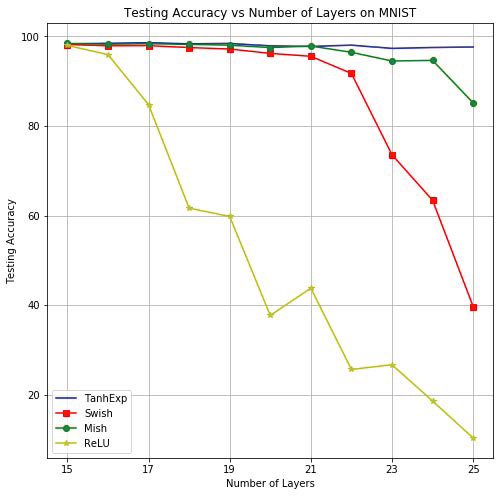}
\end{center}
   \caption{Testing accuracy (in percentage) while tuning the layers on MNIST with different activation functions. While tuning the layers, the accuracy of TanhExp remains stable.}
\label{differentlayermni}
\end{figure}

\begin{figure}[t]
\begin{center}
\includegraphics[width=0.7\linewidth]{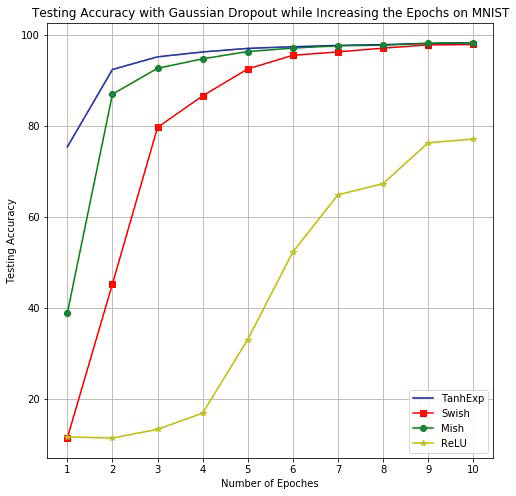}
\end{center}
   \caption{Testing accuracy (in percentage) with 15 layers on MNIST with different activation functions, with a multiplicative 1-centered Gaussian noise implemented in each layer.}
\label{gauss}
\end{figure}
\begin{figure}[t]
\begin{center}
\includegraphics[width=0.7\linewidth]{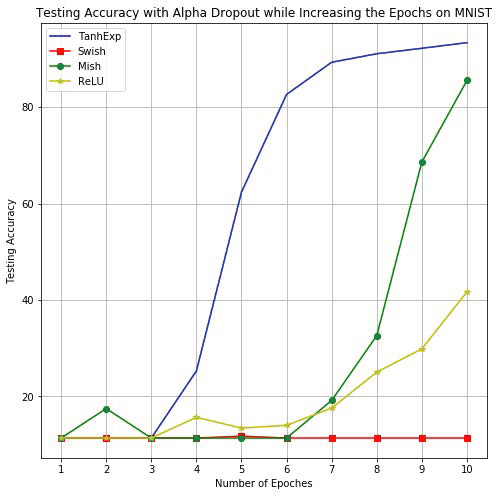}
\end{center}
   \caption{Testing accuracy (in percentage) with 15 layers on MNIST with different activation functions, with an alpha dropout replacing the dropout layer before the first dense layer.}
\label{alpha}
\end{figure}
\begin{figure}[t]
\begin{center}
\includegraphics[width=1.0\linewidth]{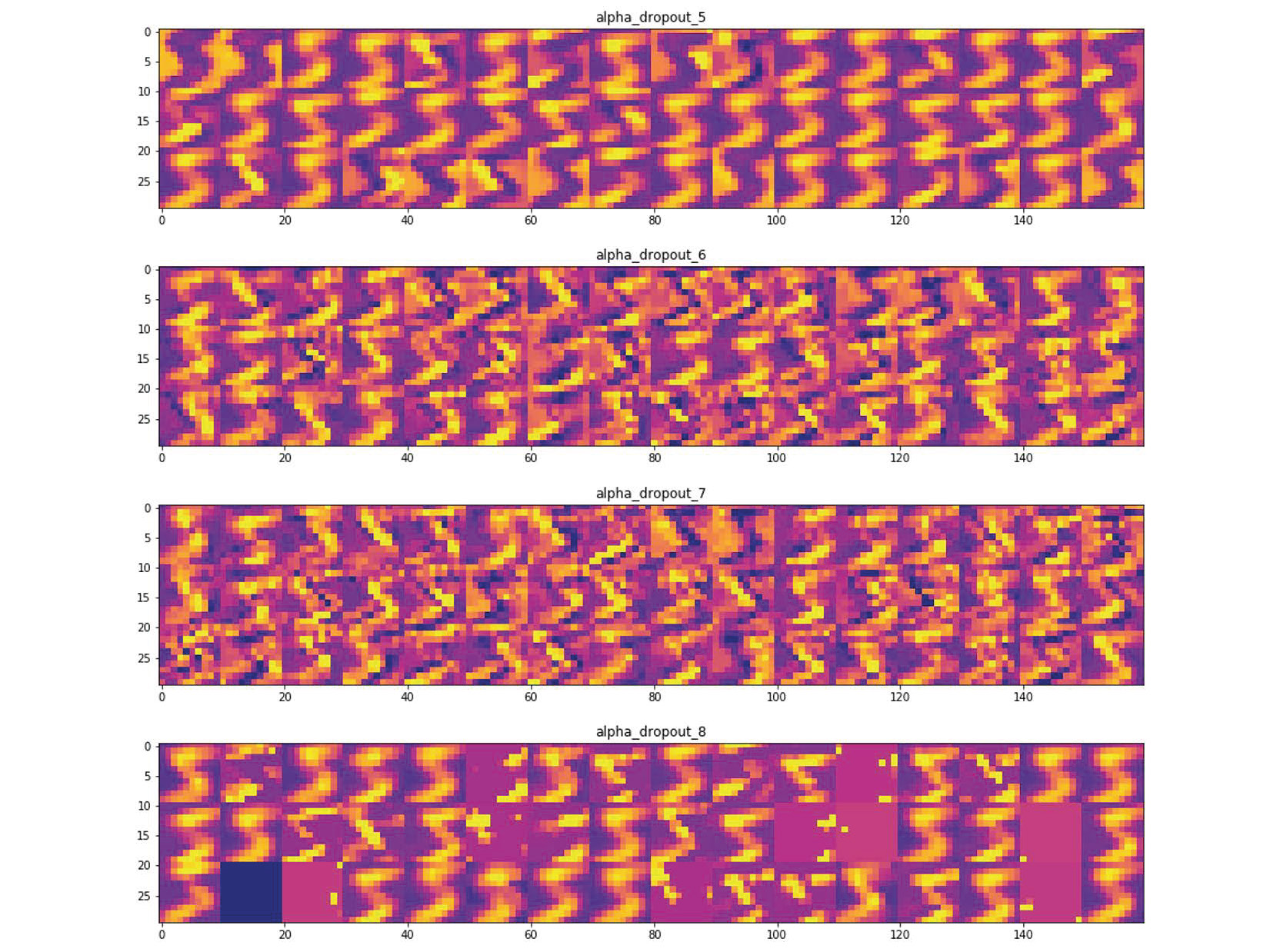}
\end{center}
   \caption{The representations of the alpha dropout layer with different activation functions. From top to bottom refers to TanhExp, Mish, Swish, and ReLU respectively. Best viewed in color.}
\label{visualize}
\end{figure}
\section{Experiments}\label{sec4}

In this section, we demostrate the properties of TanhExp in three aspects: efficiency, robustness, and simplicity. We use ReLU [7], Swish [16], Mish [6] as comparisons. For all experiments in this section, we only altered the activation functions and left all other settings unchanged at the same time.

\subsection{MNIST}

MNIST [18] is a dataset aiming at classifying the handwritten digits into 10 classes, with 60,000 training samples and 10,000 test samples. We show that TanhExp is efficient and robust to noise.

\subsubsection{Comparison of Learning Speed}

Firstly, we did the experiments with a basic network. The network architecture is illustrated in Fig. \ref{mnistnet}. The default settings contain 15 layers. We name the last 12 layers as blocks, so each block is composed of a batch normalization [19] part, an activation function, a dropout [20] rate of 0.25, and a dense layer with 500 neurons. The network is implemented following the original deisign in Mish for a fair comparison. We also tested the performance of Mish, Swish and ReLU in the same settings. With the increasing of the training epoch, the testing accuracy steadily increases while the loss decreases on all activation functions. However, in Fig. \ref{mnistacc} and Fig. \ref{mnistloss}, TanhExp outperforms Mish,  Swish, and ReLU in both the convergence speed and the final accuracy. Notice that the testing accuracy of TanhExp after the first epoch is 0.8986, while Mish is 0.6568 and Swish is 0.4030, it demonstrates that TanhExp is able to update the parameters rapidly and forces the network to fit the dataset in a more effective way, thus leads to high accuracy and low loss.

\subsubsection{Comparison of the Ability of Preventing Overfitting}

To verify that TanhExp remains stable with the number of layers growing, we varied the number of blocks from 15 to 25. Experiments were carried out with the same hyper-parameters as the above, and we visualized the final results in Fig.  \ref{differentlayermni}. Once the number of blocks reaches more than 21, ReLU and Swish show a significant decrease in accuracy, Mish also suffers from a slight decrease, while TanhExp hardly drops its accuracy, with 0.9763 at 25 layers. We assessed the results and realized that the network suffers from over-fitting when the network goes deeper. Therefore, TanhExp can prevent the network from this phenomenon markedly while other smooth activation functions do not maintain the stability as the increasing of the number of layers.

\subsubsection{Comparison of Added Noise}

To further prove the robustness of TanhExp, we implemented a multiplicative 1-centered Gaussian noise in each layer which is named Gaussian Dropout [20]. Its standard deviation is 
\begin{equation}
    Stddev = \frac{rate}{1-rate}
\end{equation}
where the drop rate remained 0.25 in this experiment. The result in Fig. \ref{gauss} illustrates that only TanhExp is barely affected by the noise. 

Another experiment is to alter the first dropout layer to an alpha dropout layer [13] with a rate of 0.2 since the network can hardly reach the global minimum value when modifying the rate to 0.25. As Fig. \ref{alpha} illustrates, alpha dropout restrains the ability of neural networks significantly, but TanhExp is still able to converge more quickly than the other activation functions. Therefore, the experiments support the statement that TanhExp is robust to added noise and has a fast convergence speed.

To validate our result in a more visualizable way, we extracted the figures of the hidden layer representations. From Fig. \ref{visualize} which is the output of the alpha dropout layer, TanhExp shows smoother and clearer representations than the other three activation functions. It is due to that the network with TanhExp as its activation function can recognize and extract key information from the entire input image more quickly, while the others with coarse representations illustrate that the network only extracts the low-level feature information partially.

Next, we explore whether the performance of TanhExp remains stable on different datasets.

\subsection{Fashion MNIST}
Fashion MNIST [21] is a dataset aiming at classifying 10 different real-world clothing classes, which consists of T-shirt, trouser, pullover, dress, coat, sandal, shirt, sneaker, bag, and ankle boot. It is similar to the original MNIST dataset [18] but with higher complexity. Fashion MNIST contains 70000 samples, with each sample is a 28*28 grayscale image.

% \begin{figure}[t]
% \begin{center}
% \includegraphics[width=1.0\linewidth]{figs/fashion.png}
% \end{center}
%   \caption{Samples in the Fashion MNIST dataset.}
% \label{fashion}
% \end{figure}

\begin{figure}[t]
\begin{center}
\includegraphics[width=0.7\linewidth]{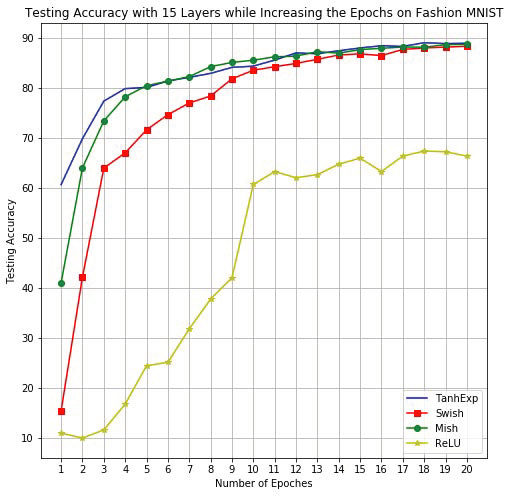}
\end{center}
   \caption{Testing accuracy (in percentage) with 15 layers on Fashion MNIST with different activation functions.}
\label{fashionacc}
\end{figure}

\begin{figure}[t]
\begin{center}
\includegraphics[width=0.7\linewidth]{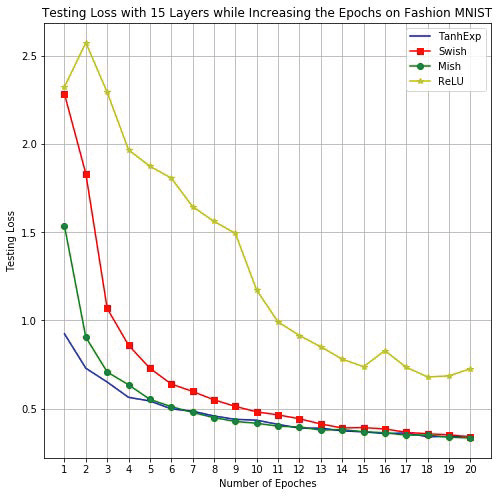}
\end{center}
   \caption{Testing loss with 15 layers on Fashion MNIST with different activation functions.}
\label{fashionloss}
\end{figure}

The model we utilized remains the same as the basic 15 layer network as in the MNIST dataset. Because the Fashion MNIST dataset is more difficult for such a simple network, we trained different models for 20 epochs each. We only changed the activation functions while tuning the model for a fair comparison. The testing accuracy and loss are illustrated in Fig. \ref{fashionacc} and Fig. \ref{fashionloss}. The results show that TanhExp outruns other activation functions evidently, on both convergence speed and eventual accuracy. 


\subsection{Kuzushiji-MNIST}

Kuzushiji-MNIST [22], also known as KMNIST, is an image classification dataset for classical Japanese literature and its 10 classes are hiragana characters. It has the same number of images as the MNIST dataset. However, the diverse distribution and the complexity of the dataset make it a more complicated task than MNIST.

% \begin{figure}[t]
% \begin{center}
% \includegraphics[width=0.8\linewidth]{figs/kmnist.png}
% \end{center}
%   \caption{Samples in the Kuzushiji-MNIST dataset.}
% \label{kmnist}
% \end{figure}

\begin{figure}[t]
\begin{center}
\includegraphics[width=0.7\linewidth]{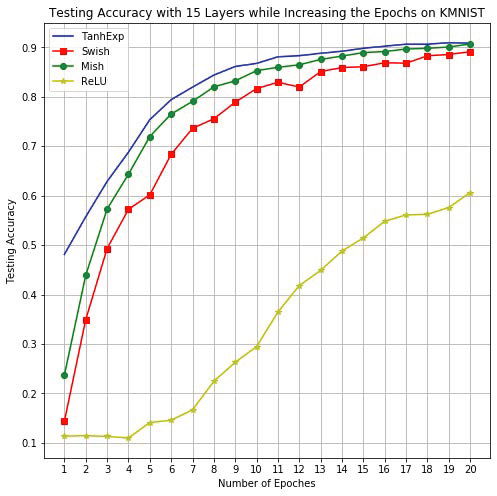}
\end{center}
   \caption{Testing accuracy (in percentage) with 15 layers on KMNIST with different activation functions.}
\label{kmnistacc}
\end{figure}

\begin{figure}[t]
\begin{center}
\includegraphics[width=0.7\linewidth]{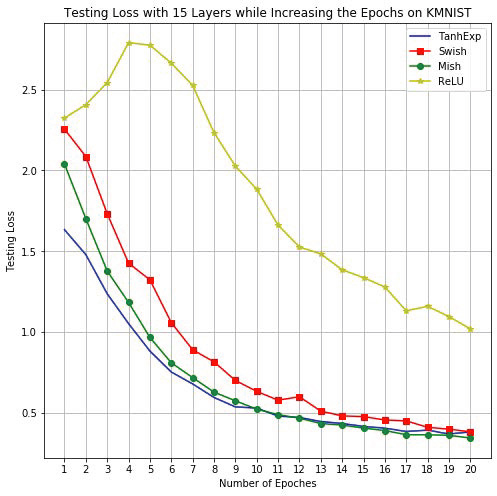}
\end{center}
   \caption{Testing loss with 15 layers on KMNIST with different activation functions.}
\label{kmnistloss}
\end{figure}

Similarly, we utilized the 15 layer network to demonstrate that TanhExp is able to perform well in various datasets and remains a high accuracy with just a few epochs. Results are shown in Figs. \ref{kmnistacc} and \ref{kmnistloss}.


\subsection{CIFAR-10}
% \begin{figure}[t]
% \begin{center}
% \includegraphics[width=0.8\linewidth]{figs/cifar10tu.png}
% \end{center}
%   \caption{Samples in the CIFAR-10 dataset.}
% \label{cifar}
% \end{figure}

\begin{table}[!b]
\begin{center}
\processtable{Testing accuracy of various models on the CIFAR-10 dataset.\label{cifar10table}}
{\begin{tabular*}{20pc}{@{\extracolsep{\fill}}lllll@{}}\toprule

Model & \textbf{TanhExp} & ReLU  & Mish & Swish\\
\midrule
LeNet [18] & \textbf{0.7262} & 0.7033  & 0.7217 & 0.7008\\
AlexNet [25] & \textbf{0.7703} & 0.7565  &  0.7626 & 0.7656\\
MobileNet [27]& 0.8538 & 0.8412 & 0.8527 & \textbf{0.8569}\\
MobileNet v2 [1]& \textbf{0.8641} & 0.8594  & 0.8605 & 0.8607\\
Resnet20 [24] & 0.9193 & 0.9150  &  0.9181 & \textbf{0.9195}\\
Resnet32 [24]& \textbf{0.9259} & 0.9178  & 0.9229 & 0.9230\\
ShuffleNet [26]& \textbf{0.8757} & 0.8705 & 0.8731 & 0.8695\\
ShuffleNet v2 [29]& \textbf{0.8743} & 0.8700 & 0.8737 & 0.8694\\
SqueezeNet [30]& \textbf{0.8852} & 0.8785 & 0.8813 & 0.8837\\
SE-Net18 [28] & \textbf{0.9086} &  0.9016 & 0.9053 & 0.8943\\
SE-Net34 [28] & 0.9119 &  \textbf{0.9167} & 0.9109 & 0.8996\\
\botrule
\end{tabular*}}{}
\end{center}
\end{table}

\begin{table}[!b]
\begin{center}
\processtable{Testing accuracy of various models on the CIFAR-100 dataset.\label{cifar100table}}
{\begin{tabular*}{20pc}{@{\extracolsep{\fill}}lllll@{}}\toprule

Model & \textbf{TanhExp} & ReLU  & Mish & Swish\\
\midrule
LeNet [18] & \textbf{0.3987} & 0.3798  & 0.3743 & 0.3809\\
AlexNet [25] & \textbf{0.4276} & 0.4191 & 0.4097 & 0.4175\\
MobileNet [27]& \textbf{0.5276} & 0.4921 & 0.5193 & 0.4995\\
MobileNet v2 [1]& \textbf{0.5737} &  0.5619 & 0.5706 & 0.5568\\
Resnet20 [24] &\textbf{0.6738}& 0.6723 &  0.6726 & 0.6710\\
Resnet32 [24] & 0.6876 & 0.6845  & \textbf{0.6944} & 0.6884\\
ShuffleNet [26]& \textbf{0.5975} & 0.5798 & 0.5919 & 0.5843\\
ShuffleNet v2 [29]& \textbf{0.6006} & 0.5856 & 0.5935 & 0.5891\\
SqueezeNet [30]& \textbf{0.6345} & 0.6093 & 0.6307 & 0.6211\\
SE-Net18 [28] & \textbf{0.6443} &  0.6272 & 0.6439 & 0.6389\\
SE-Net34 [28] & \textbf{0.6526} &  0.6456 & 0.6448 & 0.6487\\
\botrule
\end{tabular*}}{}
\end{center}
\end{table}
\begin{table}[!b]
\begin{center}
\processtable{The comparison of the computation time of different activation functions.\label{time}}
{\begin{tabular*}{20pc}{@{\extracolsep{\fill}}lllll@{}}\toprule

Function & TanhExp($\mu s$) & ReLU($\mu s$) & Mish($\mu s$)  & Swish($\mu s$) \\
\midrule
Original Function& 8.9681 & 7.0496 & 18.2508  & 6.1362 \\
1st Derivative  & 34.0548 & 10.1391 & 64.3701  &  44.9751  \\
2nd Derivative  & 56.6299 & -- & 102.2531  &  56.6815  \\

\botrule
\end{tabular*}}{}
\end{center}
\end{table}
We further explore the ability of TanhExp on other datasets that are more difficult. The images in the CIFAR-10 [23] dataset contain 10 classes, with 6000 images per class and was divided into 50000 training images and 10000 test images, each image is a 32*32 color image. These 3-channel images require a stronger learning ability of the neural network. Therefore, we used more complex lightweight neural networks in this dataset. we utilized batch normalization [19] in these different lightweight network structures [1, 24-30] with a variety of activation functions. Most experiments were carried out based on [31]. Each model was trained with a batch size of 128 and LeNet [18], AlexNet [25], Resnet20 [24], and Resnet32 [24] were trained for 200 epochs, the others for 100 epochs. The details are shown in Table \ref{cifar10table} and TanhExp surpasses ReLU, Mish, and Swish on most models and especially those tiny ones.

\subsection{CIFAR-100}
Moreover, we turned to another dataset having far more classes than the previous datasets, which is CIFAR-100 [23]. Similar to CIFAR-10, CIFAR-100 is another dataset aiming at image classification with higher complexity. It has 100 classes and each class contains 600 images which consist of 500 training images and 100 test images. We also did experiments with only modifying the activation functions and the results are shown in Table \ref{cifar100table}. The settings remained the same as CIFAR-10.

From the table, we can conclude that the TanhExp performs better than the ReLU [7], Mish [6], and Swish [16] on most of the models, with a test accuracy 5\% higher than the ReLU, 6.5\% higher than Mish, and 4.7\% higher than Swish on LeNet for instance. The results demonstrate that the proposed activation function not only convergences rapidly, but also is stable and effective, even on challenging datasets.

\subsection{Comparison of the Computation Speed}

In this subsection, we demonstrate that TanhExp not only achieves a better result in lightweight neural networks but also is simpler and requires less computation complexity. We computed TanhExp, Mish, Swish, and ReLU on a 2.20GHz Intel Xeon Cpu for $10^5$ times and get the mean computation time. We also tested their first and second derivatives. The results are shown in Table \ref{time}. The second derivative of ReLU is completely zero and calculating it is an assignment operation rather than computation, so we do not show it here. According to the table, TanhExp is about twice faster than Mish and needs approximately the same computation as Swish. However, TanhExp performs better on the most experiments with less computation time. In comparison with ReLU, TanhExp requires much fewer iterations to achieve the same accuracy as ReLU. TanhExp also have better robustness and generalization than ReLU regarding the previous experiments. Therefore, the total execution time of TanhExp is less than ReLU, meanwhile its upper-bound accuracy is higher. We can conclude that TanhExp does a better speed and accuracy trade-off than the other activation functions.

\section{Conclusion}

In this work, we propose a novel non-piecewise activation function, Tanh Exponential Activation Function, abbreviate as TanhExp, for lightweight neural networks. The equation of TanhExp is $f_{TanhExp}(x) = x\cdot \tanh(e^x)$. It is bounded below with a minimum value -0.3532 and unbounded above. The negative value could push the mean of the activations close to zero, thus accelerates the learning process. The positive part is approximately linear, with no more than 0.01 variation when the input is larger than 1, and the gradient is slightly larger than other smooth activation functions. These properties enable TanhExp to calculate and converges faster than its counterparts meanwhile provides a better result. We carried out several experiments on various datasets to demonstrate the efficiency, robustness, and simplicity of TanhExp. On MNIST, TanhExp could converge at a higher speed, with a test accuracy of 0.8986 after the first epoch on a 15-layer network, meanwhile Swish and Mish show a test accuracy of 0.4030 and 0.6568, respectively. The accuracy of TanhExp also remains stable despite the network becomes deeper. On Fashion MNIST and KMNIST, TanhExp shows the most outstanding performance in comparison with other activation functions with settings unchanged. On CIFAR-10 and CIFAR-100, TanhExp also performs well, especially on lightweight neural networks. We expect our work will promote the development of real-time manufactures. Future work will concentrate on the utilization of TanhExp on other computer vision tasks. 

\section*{Acknowledgment}

The work is partially supported by the National Natural Science Foundation of China (No. 61775048) and the Fundamental Research Funds for the Central Universities of China (No. ZDXMPY20180103).

\end{document}